%% file: WorkshopPaper.tex
\documentclass[submission]{eptcs}
\title{Interacting Conceptual Spaces}
\author{Joe Bolt \: Bob Coecke \: Fabrizio Genovese \: Martha Lewis \: Dan Marsden\: Robin Piedeleu
\footnote{Authors in alphabetical order.}
\institute{University of Oxford\\Department of Computer Science}
\email{firstname.lastname@cs.ox.ac.uk}
}

\newcommand{\event}{Semantic Spaces at the Intersection of NLP, Physics and Cognitive Science}
\def\titlerunning{Interacting Conceptual Spaces}
\def\authorrunning{J. Bolt, B. Coecke, F. Genovese, M. Lewis, D. Marsden \& R. Piedeleu}

\usepackage{amsmath,amsthm,amssymb,stmaryrd}
\usepackage{xspace}
\usepackage{subcaption}
\usepackage[center]{qtree}

\input{tikz/tikzpreamble.tex}
\newcommand{\convexrel}{\ensuremath{{\bf ConvexRel}}\xspace}

\newcommand{\lang}[1]{\ensuremath{\textit{#1}}}

\newcommand{\vect}[1]{\overrightarrow{#1}}
\newcommand{\ch}[1]{Conv({#1})}
\newcommand{\ket}[1]{\ensuremath{| #1 \rangle}}
\newcommand{\define}[1]{{\bf #1}}
\newcommand{\property}[1]{\ensuremath{\lang{p}_\lang{#1}}}

\newtheorem{thm}{Theorem}

\newtheorem{dfn}{Definition}

\theoremstyle{definition}\newtheorem{example}{Example}

\usepackage{etoolbox}
\newtoggle{COMMENTS}
\toggletrue{COMMENTS} 

\iftoggle{COMMENTS}{
  \newcommand{\margincomment}[2]{\marginpar{\tiny\color{#1}#2}}
  \newcommand{\dm}[1]{\margincomment{green}{#1}}
  \newcommand{\ml}[1]{\margincomment{red}{#1}}
  \newcommand{\bc}[1]{\margincomment{blue}{#1}}
  \newcommand{\jb}[1]{\margincomment{orange}{#1}}
}{
  \newcommand{\dm}[1]{}
  \newcommand{\ml}[1]{}
  \newcommand{\bc}[1]{}
  \newcommand{\jb}[1]{}
}

\begin{document}
\maketitle
\begin{abstract}
  We propose applying the categorical compositional scheme of \cite{CoeckeSadrzadehClark2010} to conceptual space models of cognition.
  In order to do this we introduce the category of convex relations as a new setting for categorical compositional semantics,
  emphasizing the convex structure important to conceptual space applications.
  We show how conceptual spaces for composite types such as adjectives and verbs can be constructed.
  We illustrate this new model on detailed examples.
\end{abstract}

\section{Introduction}
  How should we represent concepts and how can they be composed to form new concepts, phrases and sentences? These questions are fundamental to cognitive science.
  Conceptual spaces theory gives a way of representing structured concepts~\cite{Gardenfors2004, Gardenfors2014}, but does not provide a satisfactory account of how they compose.
  Categorical composition distributional models have provided a successful model of natural language, exploiting compositional structure in a principled manner.
  This approach works as follows.
  
  The grammatical structure of language can be formalized using Lambek's pregroup grammars~\cite{Lambek1999}, for example.
  Distributional models describe word meanings by vectors of co-occurrence statistics, derived automatically from corpus data~\cite{Lund1996}.
  The categorical compositional distributional programme unifies these two aspects in a compositional model where grammar mediates composition of meanings.
  A key insight of this approach to natural language is that both pregroups and the category
  of vector spaces carry the same abstract structure~\cite{CoeckeSadrzadehClark2010}.
  
The abstract framework of the categorical compositional scheme is actually broader in scope than natural language applications.
It can be applied in other settings in which we wish to compose meanings in a principled manner, guided by structure. 
The outline of the general programme is as follows:
\begin{enumerate}
\label{item:compstruct}
\item \begin{enumerate}
		\item Choose a compositional structure, such as a pregroup or combinatory categorial grammar.
		\item Interpret this structure as a category, the \define{grammar category}.
	\end{enumerate}
\item \begin{enumerate}
		\label{item:meaningspace}
		\item Choose or craft appropriate meaning or concept spaces, such as vector spaces, density matrices~\cite{Piedeleu2015, Bankova2016} or conceptual spaces.
		\label{item:meaningcategory}
		\item Organize these spaces into a category, the \define{semantics category}, with the same abstract structure as the grammar category.
	\end{enumerate}
\label{item:interpret}
\item Interpret the compositional structure of the grammar category in the semantics category via a functor preserving the necessary structure.
\label{item:reduction}
\item Bingo! This functor maps type reductions in the grammar category onto algorithms for composing meanings in the semantics category. 
\end{enumerate}
In order to move away from vector spaces, we construct a new categorical setting for interpreting meanings which respects the important convex structure emphasized in the conceptual spaces
literature~\cite{Gardenfors2004, Gardenfors2014}. We show that this category has the necessary abstract structure required by categorical
compositional models. We then construct convex spaces for interpreting the types for nouns, adjective and verbs.
Finally, this allows us to use the reductions of the pregroup grammar to compose meanings in conceptual spaces. We illustrate our approach with concrete examples, and go on to discuss
directions for further research, particularly in extending our approach to support the additional mathematical structure relevant to realistic models of cognition.

\section{Background}
\subsection{Pregroup Grammar}
\label{sec:pregroups}
Lambek's pregroup grammars~\cite{Lambek1999} are used to describe syntactic structure. This choice of grammar is not essential to our approach, and other forms of categorial grammar can be used~\cite{CoeckeGrefenstetteSadrzadeh2013}. A pregroup $(P,~\leq,~\cdot,~1,~(-)^l,~(-)^r)$ is a partially ordered monoid $(P, \leq, \cdot, 1)$ where each element $p\in P$ has a left adjoint $p^l$ and a right adjoint $p^r$, such that
$p^l\cdot p \leq 1 \leq p\cdot p^l$ and $p\cdot p^r \leq 1 \leq p^r \cdot p$.

For the examples in this paper, we construct our grammar from the alphabet~$\{n, s\}$ of noun and sentence types. Grammatical types such as adjectives are constructed from $n$, $s$ and their adjoints. For example, an adjective has type $n n^l$, an intransitive verb has type $n^r s$ and a transitive verb $n^r s n^l$. If a string reduces to the type $s$, the sentence is judged grammatical. The sentence \lang{Clowns tell jokes} is typed $n~(n^r s n^l)~ n$, and can be reduced to $s$ as~$n~(n^r s n^l)~ n \leq 1\cdot s n^l n \leq 1 \cdot s \cdot 1 \leq s$. If we view our pregroup as a category, these inequalities correspond to a morphism that can be expressed graphically as in equation~\ref{eq:reduction}.

\subsection{Conceptual Spaces}
\emph{Conceptual spaces} are proposed in \cite{Gardenfors2004} as a framework for representing information at the conceptual level. G\"{a}rdenfors contrasts his theory with both a symbolic, approach to concepts, and an associationist approach where concepts are represented as associations between different kinds of information elements. Instead, conceptual spaces are structures based on quality dimensions such as weight, height, hue and brightness. Concepts are roughly interpreted as convex subsets of a conceptual space. They can also have internal structure based on domains, which are sets of integrated dimensions.

Concept composition within conceptual spaces has been formalized in~\cite{AdamsRaubal2009, RickardAisbettGibbon2007, LewisLawry2016} for example. All these approaches focus on noun-noun composition, rather than utilising any more complex structure, and the way in which nouns compose often focuses on correlations between attributes in concepts. Since then, G\"ardenfors has started to formalise verb spaces, adjectives, and other linguistic structures \cite{Gardenfors2014}. However, he has not provided a systematic method for how to utilise grammatical structures within conceptual spaces. In this sense, the category-theoretic approach to concept composition we describe below will introduce a more general approach to concept composition that can apply to varying grammatical types.

\subsection{Categorical Compositional Semantics}
\label{sec:disco}
The overarching idea is to take a theory of grammar such as the pregroup grammar, and then map the grammatical structures across to whichever structure is used to provide the semantics:
\begin{align}
\label{eq:reduction}
\begin{gathered}
\input{tikz/preg-reduction.tikz}
\end{gathered}
\qquad
\mapsto
\qquad
\begin{gathered}
{%
\beginpgfgraphicnamed{String}
\InputIfFileExists{String.tikz}{}{\input{./tikz/String.tikz}}
\endpgfgraphicnamed}
\end{gathered}
\end{align}
Detailed presentations of the ideas in this section are given in~\cite{CoeckeSadrzadehClark2010,PrellerSadrzadeh2011},
and an introduction to relevant category theory is provided in~\cite{CoeckePaquette2011}.

\section{The Category of Convex Relations}
\label{sec:convexrel}
In NLP applications, meanings are typically interpreted in categories of real vector spaces.
For our intended cognitive application, we now introduce a category
that emphasizes convex structure. The familiar definition of convex set is a subset of a vector space which is closed under
forming convex combinations. In this paper we consider a more general setting that includes
convex subsets of vector spaces, but also allows us to consider some further discrete examples.

We begin with some convenient notation. For a set~$X$ we write~$\sum_i p_i \ket{ x_i }$
for a finite formal convex sum of elements of~$X$, where $p_i \in \mathbb{R}^{\geq 0}$ and~$\sum_i p_i = 1$. We then write~$D(X)$ for the set of all such sums.
Here we abuse the physicists ket notation to highlight that our sums are formal, following
a convention introduced in~\cite{Jacobs2011b}.
Equivalently, these sums can be thought of as finite probability distributions on the elements of~$X$.

A \define{convex algebra} is a set~$A$ with a function~$\alpha : D(A) \rightarrow A$ satisfying the following conditions:
\begin{equation}
  \label{eq:convexalg}
  \alpha(\ket{ a }) = a\qquad\text{ and }\qquad \alpha(\sum_{i,j} p_i q_{i,j} \ket{ a_{i,j} }) = \alpha(\sum_i p_i \ket{ \alpha(\sum_j q_{i,j} \ket{ a_{i,j} }) })
\end{equation}
Informally, $\alpha$ is a \define{mixing operation} that allows us to form convex combinations of elements,
and the equations in~\eqref{eq:convexalg} require the following good behaviour:
\begin{itemize}
\item Forming a convex combination of a single element~$a$ returns~$a$ as we would expect
\item Iterating forming convex combinations interacts as we would expect with flattening formal sums of sums
\end{itemize}
We consider some examples of convex algebras.
\begin{example}
  The closed real interval~$[0,1]$ has an obvious convex algebra structure.
  Similarly, every real or complex vector space has a natural convex algebra structure using the underlying linear structure.
\end{example}
\begin{example}[Simplices]
  For any set~$X$, the formal convex sums of elements of~$X$ themselves form a \define{free convex algebra}, which
  can also be seen as a simplex with vertices the elements of~$X$.
\end{example}
\begin{example}
  The convex space of density matrices provides another example.
\end{example}
\begin{example}
  \label{ex:fuzzy}
  For a set $X$, the functions of type $X \rightarrow [0,1]$ form a convex algebra pointwise,\linebreak
  with~$\sum_i p_i \ket{ f_i } \mapsto (\lambda x. \sum_i p_i f_i(x))$.
  We can see this as a convex algebra of fuzzy sets.
\end{example}
\begin{example}[Semilattices]
  \label{ex:semilattices}
  As a slightly less straightforward example, every affine\footnote{An affine semilattice has all finite \emph{non-empty} joins.}
  join semilattice has a convex algebra structure given by~$\sum_i p_i \ket{ a_i } = \bigvee_i a_i$.
  Notice that here the scalars~$p_i$ are discarded and play no active role. These ``discrete'' types
  of convex algebras allow us to consider objects such as the Boolean truth values.
\end{example}
\begin{example}[Trees]
  \label{ex:concretesl}
  Given a finite tree, perhaps describing some hierarchical structure, we can construct an affine semilattice in a natural way.
For example, consider a limited universe of foods, consisting of bananas, apples, and beer. Given two members of the hierarchy, their join will be the lowest level of the hierarchy which is above them both. For instance, the join of $\lang{bananas}$ and $\lang{apples}$ would be $\lang{fruit}$.
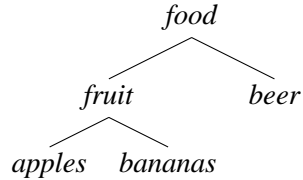
\begin{equation*}
\Tree
    [.\lang{food} [.\lang{fruit} \lang{apples} \lang{bananas} ] [.\lang{beer} ] ]
\end{equation*}
\end{example}

When~$\alpha$ can be understood from the context, we abbreviate our notation for convex combinations by writing~$\sum_i p_i a_i := \alpha(\sum_i p_i \ket{ a_i })$.
Using this convention, we define a \define{convex relation} of type\linebreak$(A,\alpha) \rightarrow (B,\beta)$ as
a binary relation~$R : A \rightarrow B$ between the underlying sets that commutes with forming convex mixtures as follows:
\begin{equation*}
  (\forall i. R(a_i, b_i)) \Rightarrow R(\sum_i p_i a_i, \sum_i p_i b_i)
\end{equation*}
We note that identity relations are convex, and convex relations are closed under relational composition and converse.
  The singleton set $\{ * \}$ has a unique convex algebra structure, denoted~$I$.
  Convex relations of the form~$I \rightarrow (A,\alpha)$ correspond to~\define{convex subsets},
  that is, subsets of~$A$ closed under forming convex combinations.
\begin{dfn}
  We define the category~\convexrel as having convex algebras as objects and convex relations
  as morphisms, with composition and identities as for ordinary binary relations.
\end{dfn}

Given a pair of convex algebras $(A,\alpha)$ and~$(B,\beta)$ we can form a new convex algebra
on the cartesian product $A \times B$, denoted~$(A,\alpha) \otimes (B,\beta)$,
with mixing operation~$\sum_i p_i \ket{ (a_i, b_i) } \mapsto (\sum_i p_i a_i, \sum_i p_i b_i)$.
This induces a symmetric monoidal structure on~\convexrel.
In fact, \convexrel has the necessary categorical structure for categorical~compositional~semantics:
\begin{thm}
  \label{thm:convexrel}
  The category~\convexrel is a dagger compact closed category 
  \footnote{We have given an elementary description of~\convexrel. More abstractly, it can be seen as the category
  of relations for the Eilenberg-Moore category of the finite distribution monad. Its compact closed
  structure then follows from general principles~\cite{CarboniWalters1987}. }.
  The symmetric monoidal structure is given by the unit and monoidal product outlined above.
  Relational converse gives a dagger structure on~\convexrel.
  The cap~$\eta_{(A,\alpha)} : I \rightarrow (A,\alpha) \otimes (A,\alpha)$ is given by~$\{ (*,(a,a)) \mid a \in A \}$ and the\linebreak
  cup~$\epsilon_{(A,\alpha)} : (A,\alpha) \otimes (A,\alpha) \rightarrow I$ is its converse.
  Every object~$(A,\alpha)$ has a canonical commutative special dagger Frobenius structure~\cite{KartsaklisSadrzadehPulmanCoecke2013},
  with copy~$\{ (a,(a,a)) \mid a \in A \} : (A,\alpha) \rightarrow (A,\alpha) \otimes (A,\alpha)$ and
  delete\linebreak
  $\{ (a,*) \mid a \in A \} : (A,\alpha) \rightarrow I$.
\end{thm}
We note that the tensor product of~\convexrel is not a category theoretic product. For example, there are
convex subsets of~$[0,1]\times[0,1]$ such as~$\{ (x,x) \mid x \in [0,1] \}$ that cannot be written as the cartesian product of two convex subsets
of~$[0,1$]. This behaviour exhibits non-trivial \emph{correlations} between the different components of the composite convex algebra.

\section{Adjective and Verb Concepts}
We define a~\define{conceptual space} to be an object of~\convexrel.
In order to match the structure of the pregroup grammar, we require two distinct objects: a noun space~$N$ and a sentence space~$S$.

The \define{noun~space}~$N$ is given by a composite~$N_\lang{colour} \otimes N_\lang{taste} \otimes ...$ describing different attributes such as colour and taste.
A~\define{noun} is then a convex subset of such a space. 
A sentence space is a convex algebra in which individual points are events. A~\define{sentence} is then a convex subset of~$S$.

We now describe some example noun and sentence spaces. We then show how these can be combined to form spaces describing adjectives and verbs.
Once we have these types available, we show in section~\ref{sec:composing} how concepts interact within sentences.

\subsection{Example: Food and Drink}
We consider a conceptual space for food and drink as our running example. The space~$N$ is composed of the domains $N_\lang{colour}$, $N_\lang{taste}$, $N_\lang{texture}$, so that $N = N_\lang{colour} \otimes N_\lang{taste} \otimes N_\lang{texture}$. The domain~$N_\lang{colour}$ is the HSV colour domain, i.e. triples $(H, S, V)$ such that $H \in[0, 360]$, $S\in[0, 1]$, $V\in[0,1]$. $N_\lang{taste}$ consists of 5-tuples $\vec{t}$ for the dimensions sweet, sour, bitter, salt and savoury, with each dimension taking values on $[0,1]$. $N_\lang{texture}$ is just the set $[0, 1]$ ranging from completely liquid (0) to completely solid (1).
We define a property \property{property} to be a convex subset of a domain, and specify the following examples:

\begin{gather*}
\property{yellow} = [45, 75] \times [0.5, 1] \times [0, 1],~\property{green} = [75, 135] \times [0.5, 1] \times [0, 1] \\
\property{brown} = [0, 45] \times [0.8, 1] \times [0.2, 0.4],~
\property{sweet} = \{\vec{t}| t_\lang{sweet} \geq 0.5 \text{ and } t_l \leq 0.5 \text{ for } l \neq \lang{sweet} \}
\end{gather*}

The properties~$\property{sour}$ and~$\property{bitter}$ are defined analogously.
We define some nouns below, using~$\ch{A}$ to denote the convex hull of a set A.
\begin{align*}
\lang{banana} &=  [60, 95] \times [0.75, 1] \times [0.25, 1] \times \ch{\property{sweet} \cup \property{bitter}} \times [0.2, 0.5]\\
\lang{apple} &= [0, 105] \times [0.75, 1] \times [0.5, 1] \times \ch{\property{sweet} \cup \property{sour}} \times [0.5, 0.8]\\
\lang{beer} &= [40, 50] \times [0.85, 1] \times [0.1, 0.7] \times \ch{\property{sweet} \cup \property{sour} \cup \property{bitter}} \times [0, 0.01]
\end{align*}

What should the sentence space for food and drink be like? We need to describe the events associated with eating and drinking. We give a very simple example where the events are either positive or negative, and surprising or unsurprising. We therefore use a sentence space of 2-tuples. The first element of the tuple states whether the sentence is positive (1) or negative (0) and the second states whether it is surprising (1) or unsurprising (0). The convex structure on this space is the convex algebra on a join semilattice induced by element-wise max, as in example \ref{ex:semilattices}. We therefore have four points in the space: positive, surprising (1,~1); positive, unsurprising (1,~0); negative, surprising (0,~1); and negative, unsurprising (0,~0). Sentence meanings are convex subsets of this space, so they could be singletons, or larger subsets such as $\lang{negative} = \{(0,~1), (0,~0)\}$.

\subsection{Adjectives}
Recall that in a pregroup setting the adjective type is $n n^l$. In \convexrel, the adjective therefore has type $N\otimes N$.
Adjectives are convex relations on the noun space, so can be written as sets of ordered pairs. We give two examples, $\lang{yellow}_\lang{adj}$ and $\lang{soft}_\lang{adj}$.
The adjective~$\lang{yellow}_\lang{adj}$ has the simple form\linebreak
$\{(\vect{x}, \vect{x}) | x_\lang{colour} \in \property{yellow} \}$
because it depends only on one area of the conceptual space.

An adjective such as `soft' behaves differently to this. We cannot simply define soft as one area of the conceptual space, because whether or not something is soft depends what it was originally. Using relations, we can start to write down the right type of structure for the adjective, as long as the objects are sufficiently distinct. Restricting our universe just to bananas and apples, we can write~$\lang{soft}_\lang{adj}$\linebreak
as~$\{(\vect{x}, \vect{x}) | \vect{x} \in \lang{banana} \text{ and } x_\lang{texture} \leq 0.35 \text{ or } \vect{x} \in \lang{apples} \text{ and } x_\lang{texture} \leq 0.6\}$

An analysis of the difficulties in dealing with adjectives set-theoretically, breaking them down into (roughly) three categories, is given in~\cite{KampPartee1995}.
Under this view, both adjectives and nouns are viewed as one-place predicates, so that, for example~$\lang{red} = \{ x | \text{$x$ is red}\}$ and~$\lang{dog} = \{ x |\text{$x$ is a dog}\}$.
There are then three classes of adjective. For \define{intersective} adjectives,  the meaning of $\lang{adj noun}$ is given by  $\lang{adj} \cap \lang{noun}$. For \define{subsective} adjectives, the meaning of $\lang{adj noun}$ is a subset of  $\lang{noun}$. For \define{privative} adjectives, however, $\lang{adj noun} \not\subseteq \lang{noun}$. 

Intersective adjectives are a simple modifier that can be thought of as the intersection between two concepts.
We can make explicit the internal structure of these adjectives exploiting the Frobenius structure of theorem~\ref{thm:convexrel}.
For example, in the case of $\lang{yellow banana}$, we take the intersection of $\lang{yellow}$ and $\lang{banana}$.
Using the Frobenius axioms, we then show how to understand~$\lang{yellow}$ as an adjective. Below, we show the general case of an adjective to the left, and an intersective adjective to the right.
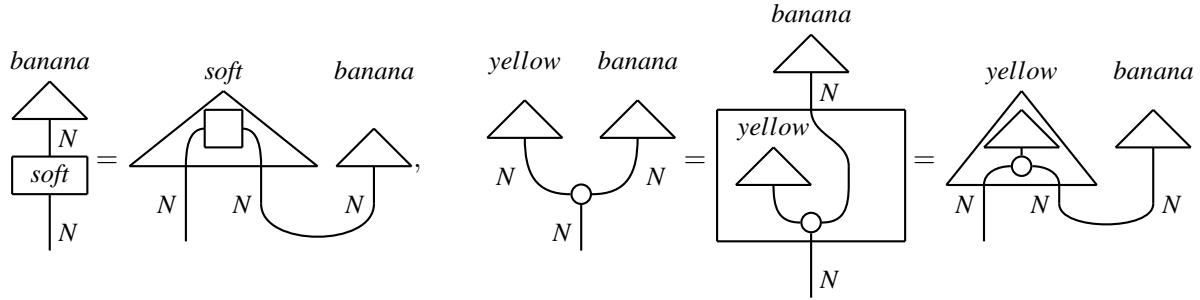
\begin{align*}
\begin{gathered} \input{tikz/ftall.tikz}\end{gathered} = \begin{gathered} \input{tikz/tall_bent.tikz}\end{gathered},
\qquad
\begin{gathered} \input{tikz/yellow_int.tikz}\end{gathered} = \begin{gathered} \input{tikz/yellow_adj.tikz}\end{gathered}= \begin{gathered} \input{tikz/yellow_bent.tikz}\end{gathered}
\end{align*}
This shows us how the internal structure of an intersective adjective is derived directly from a noun.

\subsection{Verbs}
The pregroup type for a transitive verb is $n^r s n^l$, mapping to $N \otimes S\otimes N$ in \convexrel. To define the verb, we use concept names as shorthand, where these can easily be calculated. For example, 
\begin{align*}
\lang{green banana} &= [75, 95] \times [0.75, 1] \times [0.25, 1] \times \ch{\property{sweet} \cup \property{bitter}} \times [0.2, 0.5]\\
\lang{bitter} &= N_\lang{colour} \times \property{bitter} \times N_\lang{texture}
\end{align*}
Although a full specification of a verb would take in all the nouns it could possibly apply to, for expository purposes we restrict our nouns to just bananas and beer which do not overlap, due to the fact that they have different textures. We define the verb $\lang{taste} : I \rightarrow N \otimes S \otimes N$ as follows:
\begin{align*}
\lang{taste} &= (\lang{green banana} \times \{(0, 0)\} \times \lang{bitter}) \cup  (\lang{green banana} \times \{(1, 1)\} \times \lang{sweet})\\
&\quad  \cup (\lang{yellow banana} \times \{(1, 0)\} \times \lang{sweet}) \\
&\quad \cup (\lang{beer} \times \{(0, 1)\}  \times \lang{sweet}) \cup (\lang{beer} \times \{(1, 0)\} \times \lang{bitter})
\end{align*}

\section{Concepts in Interaction}
\label{sec:composing}
We have given descriptions of how to form the different word types within our model of categorical conceptual spaces. In this section we show how to apply the type reductions of the pregroup grammar within the conceptual spaces formalism.
The application of $\lang{yellow}_\lang{adj}$ to \lang{banana} works as follows.
\begin{align*}
\lang{yellow banana} &= (1_N \times \epsilon_N)(\lang{yellow}_\lang{adj} \times \lang{banana})\\
&=(1_N \times \epsilon_N)\{(\vect{x}, \vect{x}) | x_\lang{colour} \in \lang{yellow} \}\\
&\qquad \times ([60, 95] \times [0.75, 1] \times [0.25, 1] \times \ch{\property{sweet} \cup \property{bitter}} \times [0.2, 0.5])\\
&= [60, 75] \times [0.75, 1] \times [0.25, 1] \times \ch{\property{sweet} \cup \property{bitter}} \times [0.2, 0.5]
\end{align*}
Notice, in the last line, how the hue element of the colour domain has altered from $[60, 95]$ to $[60, 75]$.
This assumes that we can tell bananas and apples apart by shape, colour and so on. Then the same calculation gives us $\lang{soft apple} =  [0, 105] \times [0.75, 1] \times [0.5, 1] \times \ch{\property{sweet} \cup \property{sour}} \times [0.4, 0.6]$.

Using the definition of \lang{taste} that we gave, we find that although sweet bananas are good:
\begin{align*}
&\lang{bananas taste sweet} = (\epsilon_N \times 1_S \times \epsilon_N)(\lang{bananas}\times\lang{taste}\times\lang{sweet})\\
&\qquad = (\epsilon_N \times 1_S)(\lang{banana} \times (\lang{green banana} \times \{(1, 1)\} \cup \lang{yellow banana} \times \{(1, 0)\})\\\nonumber
&\qquad=\{(1,1), (1, 0)\} = \lang{positive}
\end{align*}
sweet beer is not so desirable:
\begin{equation*}
\lang{beer tastes sweet} = (\epsilon_N \times 1_S \times \epsilon_N)(\lang{beer}\times\lang{taste}\times\lang{sweet}) = \{(0, 1)\} = \lang{negative and surprising}
\end{equation*}
\paragraph{Relative Pronouns}
The compositional semantics we use can also deal with relative pronouns, described in detail in \cite{KartsaklisSadrzadehPulmanCoecke2013}. Relative pronouns are words such as `which'. For example, we can form the noun phrase \lang{Fruit which tastes bitter}. This has the structure given in equation \ref{eq:frobenius}:
\begin{align}
\label{eq:frobenius}
\begin{gathered}
\input{tikz/frob-sub.tikz}
\end{gathered}
\quad
=
\quad
\begin{gathered}
\input{tikz/frob-yanked.tikz}
\end{gathered}
\end{align}
In our example, we find that $\lang{Fruit which tastes bitter} = \lang{green banana}$:
\begin{align*}
\lang{Fruit which tastes bitter} &= (\mu_N \times \iota_S \times \epsilon_N)(\ch{\lang{bananas} \cup \lang{apples}} \times \lang{taste} \times \lang{bitter})\\
&=(\mu_N\times \iota_S)(\ch{\lang{bananas} \cup \lang{apples}} \times (\lang{green banana} \times \{(0, 0)\}))\\
&=\mu_N(\ch{\lang{bananas} \cup \lang{apples}} \times (\lang{green banana})) = \lang{green banana}
\end{align*}
where $\mu_N$ is the converse of the Frobenius copy map on $N$ and $\iota_S$ is the delete map on $S$ from theorem \ref{thm:convexrel}.
\section{Conclusion}
We have applied the categorical compositional scheme
to cognition and conceptual spaces. In order to do this we introduced a new model for categorical
compositional semantics, the category~\convexrel of convex algebras and binary relations respecting
convex structure. We consider this model as a proof of concept that we can describe convex structures within
our framework. Conceptual spaces are often considered to have further mathematical structure such as distance
measures and notions of convergence or fixed points. It is also possible to vary the notion of convexity under
consideration, for example by considering a binary betweenness relation on a space as primitive, rather than
a mixing operation. Identifying a good setting for rich conceptual spaces models, and incorporating those
structures into a compositional framework is a direction for further work. In particular, ongoing work includes developing a notion of negation that depends on reversing the natural ordering on concepts that arises via subset inclusion.


\subsubsection*{Acknowledgements}
This work was partially funded by AFSOR grant ``Algorithmic and Logical Aspects when Composing Meanings'', the FQXi grant ``Categorical Compositional Physics'', and EPSRC.

\bibliographystyle{eptcs}
\bibliography{cs}
\end{document}

%% file: tikz/tikzpreamble.tex
\usepackage{tikz}
\usepackage{pgfplots}
\usetikzlibrary{trees}
\usetikzlibrary{topaths}
\usetikzlibrary{decorations.pathmorphing}
\usetikzlibrary{decorations.markings}
\usetikzlibrary{matrix,backgrounds,folding}
\usetikzlibrary{chains,scopes,positioning,fit}
\usetikzlibrary{arrows,shadows}
\usetikzlibrary{calc} 
\usetikzlibrary{chains}
\usetikzlibrary{shapes,shapes.geometric,shapes.misc}

\pgfdeclarelayer{edgelayer}
\pgfdeclarelayer{nodelayer}
\pgfsetlayers{background,edgelayer,nodelayer,main}

\tikzstyle{dot}=[circle,fill=black,draw=black]
\tikzstyle{none}=[inner sep=0pt]
\tikzstyle{every loop}=[]

\tikzstyle{(null)}=[]
\tikzstyle{plain}=[]

\tikzstyle{blank}=[inner sep=0pt]
\tikzstyle{box}=[rectangle, minimum size = 0.5cm,fill=white,draw=black]
\tikzstyle{small_node}=[thick, inner sep=0pt, minimum size=0.2cm,circle,fill=white,draw=black]
\tikzstyle{square}=[thick, minimum size=0.3cm,rectangle,fill=white,draw=black]

\tikzstyle{downtri}=[regular polygon,regular polygon sides=3,shape border rotate=180,fill=white,draw=black,line width=0.8 pt]
\tikzstyle{uptri}=[regular polygon,regular polygon sides=3,shape border rotate=0,fill=white,draw=black]

\tikzstyle{morph}=[->,draw=black,line width=0.600]
\tikzstyle{arrow}=[-,draw=black,postaction={decorate},decoration={markings,mark=at position .5 with {\arrow{>}}},line width=2.000]
\tikzstyle{tick}=[-,draw=black,postaction={decorate},decoration={markings,mark=at position .5 with {\draw (0,-0.1) -- (0,0.1);}},line width=2.000]

%% file: tikz/preg-reduction.tikz
\begin{tikzpicture}[text height=1.5 ex]
	\begin{pgfonlayer}{nodelayer}
		\node [style=none] (3) at (-1.75, 3) {$n$};
		\node [style=none] (4) at (0, 3) {$s$};
		\node [style=none] (5) at (1.75, 3) {$n$};
		\node [style=none] (6) at (-0.5, 3) {$n^r$};
		\node [style=none] (7) at (0.5, 3) {$n^l$};
		\node [style=none] (8) at (-1.75, 2.75) {};
		\node [style=none] (9) at (-0.5, 2.75) {};
		\node [style=none] (10) at (0, 2.75) {};
		\node [style=none] (11) at (0.5, 2.75) {};
		\node [style=none] (12) at (1.75, 2.75) {};
		\node [style=none] (13) at (0, 2) {};
	\end{pgfonlayer}
	\begin{pgfonlayer}{edgelayer}
		\draw [thick, bend right=90, looseness=1.00] (8.center) to (9.center);
		\draw [thick, bend right=90, looseness=1.00] (11.center) to (12.center);
		\draw [thick] (10.center) to (13.center);
	\end{pgfonlayer}
\end{tikzpicture}

%% file: tikz/String.tikz
\begin{tikzpicture}[scale=0.8, text height=1.5 ex]
	\begin{pgfonlayer}{nodelayer}
		\node [style=none] (0) at (-2.25, 0) {};
		\node [style=none] (1) at (-1.25, 0) {};
		\node [style=none] (2) at (-1.75, 0.5) {};
		\node [style=none] (3) at (-1, 0) {};
		\node [style=none] (4) at (1, 0) {};
		\node [style=none] (5) at (0, 0.75) {};
		\node [style=none] (6) at (2.25, 0) {};
		\node [style=none] (7) at (1.75, 0.5) {};
		\node [style=none] (8) at (1.25, 0) {};
		\node [style=none] (9) at (-1.75, 0) {};
		\node [style=none] (10) at (-0.5, 0) {};
		\node [style=none] (11) at (0, 0) {};
		\node [style=none] (12) at (0.5, 0) {};
		\node [style=none] (13) at (1.75, 0) {};
		\node [style=none] (14) at (-1.75, -0.25) {};
		\node [style=none] (15) at (-0.5, -0.25) {};
		\node [style=none] (16) at (0, -0.25) {};
		\node [style=none] (17) at (0.5, -0.25) {};
		\node [style=none] (18) at (1.75, -0.25) {};
		\node [style=none, anchor=base, yshift=-1.2mm] (19) at (-1.75, -0.5) {\small $N$};
		\node [style=none, anchor=base, yshift=-1.2mm] (20) at (-0.5, -0.5) {\small $N$};
		\node [style=none, anchor=base, yshift=-1.2mm] (21) at (0, -0.5) {\small $S$};
		\node [style=none, anchor=base, yshift=-1.2mm] (22) at (0.5, -0.5)  {\small $N$};
		\node [style=none, anchor=base, yshift=-1.2mm] (23) at (1.75, -0.5) {\small $N$};
		\node [style=none] (24) at (-1.75, -0.75) {};
		\node [style=none] (25) at (-0.5, -0.75) {};
		\node [style=none] (26) at (0, -0.75) {};
		\node [style=none] (27) at (0.5, -0.75) {};
		\node [style=none] (28) at (1.75, -0.75) {};
		\node [style=none] (29) at (0, -1.5) {};
		\node [style=none, anchor=base, yshift=-1.2mm] (30) at (-1.75, 1) {\small Clowns};
		\node [style=none, anchor=base, yshift=-1.2mm] (31) at (0, 1) {\small tell};
		\node [style=none, anchor=base, yshift=-1.2mm] (32) at (1.75, 1) {\small jokes};
	\end{pgfonlayer}
	\begin{pgfonlayer}{edgelayer}
		\draw [style=thick] (0.center) to (1.center);
		\draw [style=thick] (1.center) to (2.center);
		\draw [style=thick] (2.center) to (0.center);
		\draw [style=thick] (3.center) to (4.center);
		\draw [style=thick] (8.center) to (6.center);
		\draw [style=thick] (6.center) to (7.center);
		\draw [style=thick] (4.center) to (5.center);
		\draw [style=thick] (3.center) to (5.center);
		\draw [style=thick] (8.center) to (7.center);
		\draw [style=thick] (9.center) to (14.center);
		\draw [style=thick] (10.center) to (15.center);
		\draw [style=thick] (11.center) to (16.center);
		\draw [style=thick] (12.center) to (17.center);
		\draw [style=thick] (13.center) to (18.center);
		\draw [thick, bend right=90, looseness=1.25] (24.center) to (25.center);
		\draw [thick, bend right=90, looseness=1.25] (27.center) to (28.center);
		\draw [style=thick] (26.center) to (29.center);
	\end{pgfonlayer}
\end{tikzpicture}

%% file: tikz/ftall.tikz
\begin{tikzpicture}[text height=1.5 ex]
	\begin{pgfonlayer}{nodelayer}
		\node [style=none] (0) at (-0.5, 1.25) {};
		\node [style=none] (1) at (0.5, 1.25) {};
		\node [style=none] (2) at (0, 1.75) {};
		\node [style=none] (3) at (0, 1.25) {};
		\node [style=none] (4) at (0, 0.75) {};
		\node [style=none] (5) at (0, 0.25) {};
		\node [style=none] (6) at (0, -0.5) {};
		\node [style=none] (7) at (-0.5, 0.75) {};
		\node [style=none] (8) at (-0.5, 0.25) {};
		\node [style=none] (9) at (0.5, 0.25) {};
		\node [style=none] (10) at (0.5, 0.75) {};
		\node [style=none, anchor=mid] (11) at (0, 0.5) {\small$\lang{soft}$};
		\node [style=none] (12) at (0.25, 1) {\small$N$};
		\node [style=none] (13) at (0.25, -0.25) {\small$N$};
		\node [style=none, anchor=mid] (14) at (0, 2) {\small$\lang{banana}$};
	\end{pgfonlayer}
	\begin{pgfonlayer}{edgelayer}
		\draw [thick] (0.center) to (1.center);
		\draw [thick] (1.center) to (2.center);
		\draw [thick] (2.center) to (0.center);
		\draw [thick] (3.center) to (4.center);
		\draw [thick] (5.center) to (6.center);
		\draw [thick] (7.center) to (8.center);
		\draw [thick] (8.center) to (9.center);
		\draw [thick] (9.center) to (10.center);
		\draw [thick] (10.center) to (7.center);
	\end{pgfonlayer}
\end{tikzpicture}

%% file: tikz/tall_bent.tikz
\begin{tikzpicture}[text height=1.5 ex]
	\begin{pgfonlayer}{nodelayer}
		\node [style=none] (0) at (0.5, -0.25) {};
		\node [style=none] (1) at (1, -0.25) {};
		\node [style=none] (2) at (-1.25, 0.5) {};
		\node [style=none] (3) at (-1.25, 0) {};
		\node [style=none, anchor=mid] (4) at (-1.5, 1) {\small$\lang{soft}$};
		\node [style=none] (5) at (0.5, -0.75) {};
		\node [style=none] (6) at (-2, -0.25) {};
		\node [style=none] (7) at (0.5, 0.25) {};
		\node [style=none] (8) at (-2, -1.25) {};
		\node [style=none] (9) at (-1.75, 0.5) {};
		\node [style=none] (10) at (0, -0.25) {};
		\node [style=none] (11) at (-0.25, -0.25) {};
		\node [style=none] (12) at (-1.5, 0.75) {};
		\node [style=none] (13) at (-2.75, -0.25) {};
		\node [style=none] (14) at (-1.75, 0.25) {};
		\node [style=none] (15) at (-1.25, 0.25) {};
		\node [style=none] (16) at (-1, -0.25) {};
		\node [style=none] (17) at (-1.75, 0) {};
		\node [style=none] (18) at (0.25, -0.75) {\small$N$};
		\node [style=none] (19) at (-1.25, -0.75) {\small$N$};
		\node [style=none] (20) at (-2.25, -0.75) {\small$N$};
		\node [style=none] (21) at (-1, -0.75) {};
		\node [style=none, anchor=mid] (22) at (0.5, 1) {\small$\lang{banana}$};
	\end{pgfonlayer}
	\begin{pgfonlayer}{edgelayer}
		\draw [thick] (1.center) to (10.center);
		\draw [thick] (10.center) to (7.center);
		\draw [thick] (7.center) to (1.center);
		\draw [thick] (0.center) to (5.center);
		\draw [thick] (6.center) to (8.center);
		\draw [thick] (2.center) to (3.center);
		\draw [thick] (9.center) to (2.center);
		\draw [thick, in=0, out=90, looseness=1.00] (16.center) to (15.center);
		\draw [thick, in=90, out=180, looseness=1.00] (14.center) to (6.center);
		\draw [thick] (11.center) to (12.center);
		\draw [thick] (12.center) to (13.center);
		\draw [thick] (11.center) to (13.center);
		\draw [thick] (17.center) to (3.center);
		\draw [thick] (17.center) to (9.center);
		\draw [thick] (16.center) to (21.center);
		\draw [thick, bend right=90, looseness=1.00] (21.center) to (5.center);
	\end{pgfonlayer}
\end{tikzpicture}

%% file: tikz/yellow_int.tikz
\begin{tikzpicture}[text height=1.5 ex]
	\begin{pgfonlayer}{nodelayer}
		\node [style=none] (0) at (-0.25, 0) {};
		\node [style=none] (1) at (-1.25, 0) {};
		\node [style=none] (2) at (0.75, 0.5) {};
		\node [style=none] (3) at (1.25, 0) {};
		\node [style=none] (4) at (0.25, 0) {};
		\node [style=none] (5) at (-0.75, 0) {};
		\node [style=none, anchor=mid] (6) at (-0.75, 1) {\small$yellow$};
		\node [style={small_node}] (7) at (0, -0.75) {};
		\node [style=none] (8) at (0.75, 0) {};
		\node [style=none] (9) at (-0.75, 0.5) {};
		\node [style=none, anchor=mid] (10) at (0.75, 1) {\small$banana$};
		\node [style=none] (11) at (-0.25, -1.25) {\small$N$};
		\node [style=none] (12) at (1, -0.5) {\small$N$};
		\node [style=none] (13) at (0, -1.5) {};
		\node [style=none] (14) at (-1, -0.5) {\small$N$};
	\end{pgfonlayer}
	\begin{pgfonlayer}{edgelayer}
		\draw [thick] (4.center) to (3.center);
		\draw [thick] (3.center) to (2.center);
		\draw [thick] (2.center) to (4.center);
		\draw [thick] (9.center) to (1.center);
		\draw [thick] (1.center) to (0.center);
		\draw [thick] (0.center) to (9.center);
		\draw [thick, in=180, out=-90, looseness=1.25] (5.center) to (7);
		\draw [thick] (7) to (13.center);
		\draw [thick, in=0, out=-90, looseness=1.25] (8.center) to (7);
	\end{pgfonlayer}
\end{tikzpicture}

%% file: tikz/yellow_adj.tikz
\begin{tikzpicture}[text height=1.5 ex]
	\begin{pgfonlayer}{nodelayer}
		\node [style=none] (0) at (-0.5, 1.5) {};
		\node [style=none] (1) at (0.5, 1.5) {};
		\node [style=none] (2) at (0, 2) {};
		\node [style=none] (3) at (0, 1.5) {};
		\node [style=none] (4) at (0.5, 0.25) {};
		\node [style=none] (5) at (0, -0.75) {};
		\node [style=none] (6) at (0, -1.5) {};
		\node [style=none] (7) at (-1.25, 1) {};
		\node [style=none] (8) at (-1.25, -0.75) {};
		\node [style=none] (9) at (1.25, -0.75) {};
		\node [style=none] (10) at (1.25, 1) {};
		\node [style=none, anchor=mid] (11) at (-0.5, 0.75) {\small$yellow$};
		\node [style=none] (12) at (0.25, 1.25) {\small$N$};
		\node [style=none] (13) at (0.25, -1.25) {\small$N$};
		\node [style=none] (14) at (-0.5, 0.5) {};
		\node [style=none] (15) at (-1, 0) {};
		\node [style=none] (16) at (0, 0) {};
		\node [style=none] (17) at (-0.5, 0) {};
		\node [style={small_node}] (18) at (0, -0.5) {};
		\node [style=none] (19) at (0.5, 0) {};
		\node [style=none] (20) at (0, 1) {};
		\node [style=none, anchor=mid] (21) at (0, 2.25) {\small$banana$};
	\end{pgfonlayer}
	\begin{pgfonlayer}{edgelayer}
		\draw [thick] (0.center) to (1.center);
		\draw [thick] (1.center) to (2.center);
		\draw [thick] (2.center) to (0.center);
		\draw [thick] (5.center) to (6.center);
		\draw [thick] (7.center) to (8.center);
		\draw [thick] (8.center) to (9.center);
		\draw [thick] (9.center) to (10.center);
		\draw [thick] (10.center) to (7.center);
		\draw [thick] (14.center) to (15.center);
		\draw [thick] (15.center) to (16.center);
		\draw [thick] (16.center) to (14.center);
		\draw [thick, in=180, out=-90, looseness=1.25] (17.center) to (18);
		\draw [thick] (18) to (5.center);
		\draw [thick] (4.center) to (19.center);
		\draw [thick, in=0, out=-90, looseness=1.25] (19.center) to (18);
		\draw [thick] (3.center) to (20.center);
		\draw [thick, in=90, out=-90, looseness=1.00] (20.center) to (4.center);
	\end{pgfonlayer}
\end{tikzpicture}

%% file: tikz/yellow_bent.tikz
\begin{tikzpicture}[text height=1.5 ex]
	\begin{pgfonlayer}{nodelayer}
		\node [style=none] (0) at (0.5, 0) {};
		\node [style=none] (1) at (-0.5, 0) {};
		\node [style=none] (2) at (1.75, 0.5) {};
		\node [style=none] (3) at (2.25, 0) {};
		\node [style=none] (4) at (1.25, 0) {};
		\node [style=none] (5) at (0, 0) {};
		\node [style=none,anchor=mid] (6) at (0, 1) {\small$yellow$};
		\node [style={small_node}] (7) at (0, -0.25) {};
		\node [style=none] (8) at (1.75, 0) {};
		\node [style=none] (9) at (0, 0.5) {};
		\node [style=none, anchor=mid] (10) at (1.75, 1) {\small$banana$};
		\node [style=none] (11) at (-0.5, -0.5) {};
		\node [style=none] (12) at (1.75, -0.75) {};
		\node [style=none] (13) at (-0.75, -0.75) {\small$N$};
		\node [style=none] (14) at (2, -0.75) {\small$N$};
		\node [style=none] (15) at (0.5, -0.5) {};
		\node [style=none] (16) at (0, 0.75) {};
		\node [style=none] (17) at (-1, -0.5) {};
		\node [style=none] (18) at (1, -0.5) {};
		\node [style=none] (19) at (-0.5, -1.25) {};
		\node [style=none] (20) at (0.5, -0.75) {};
		\node [style=none] (21) at (0.25, -0.75) {\small$N$};
	\end{pgfonlayer}
	\begin{pgfonlayer}{edgelayer}
		\draw [thick] (4.center) to (3.center);
		\draw [thick] (3.center) to (2.center);
		\draw [thick] (2.center) to (4.center);
		\draw [thick] (9.center) to (1.center);
		\draw [thick] (1.center) to (0.center);
		\draw [thick] (0.center) to (9.center);
		\draw [thick, in=90, out=-90, looseness=1.00] (5.center) to (7);
		\draw [thick] (8.center) to (12.center);
		\draw [thick, in=90, out=180, looseness=1.00] (7) to (11.center);
		\draw [thick, in=90, out=0, looseness=1.00] (7) to (15.center);
		\draw [thick] (16.center) to (17.center);
		\draw [thick] (17.center) to (18.center);
		\draw [thick] (16.center) to (18.center);
		\draw [thick] (11.center) to (19.center);
		\draw [thick] (15.center) to (20.center);
		\draw [thick, bend right=90, looseness=0.75] (20.center) to (12.center);
	\end{pgfonlayer}
\end{tikzpicture}

%% file: tikz/frob-sub.tikz
\begin{tikzpicture}[scale=0.55]
	\begin{pgfonlayer}{nodelayer}
		\node [style=none] (0) at (-3.5, 1.25) {\lang{Fruit}};
		\node [style=none] (1) at (-1, 1.25) {\lang{which}};
		\node [style=none] (2) at (1.75, 1.25) {\lang{tastes}};
		\node [style=none] (3) at (3.75, 1.25) {\lang{bitter}};
		\node [style=none] (4) at (-3.5, -0.5) {};
		\node [style=none] (5) at (-2.5, -0.5) {};
		\node [style=none] (6) at (-1.25, -0.5) {};
		\node [style=none] (7) at (0.25, -0.5) {};
		\node [style=none] (8) at (1.25, -0.5) {};
		\node [style=none] (9) at (1.75, -0.5) {};
		\node [style=none] (10) at (2.25, -0.5) {};
		\node [style=none] (11) at (3.75, -0.5) {};
		\node [style={small_node}] (12) at (-1.25, -0.25) {};
		\node [style=none] (13) at (-2.5, 0.25) {};
		\node [style=none] (14) at (-1.75, 0.25) {};
		\node [style=none] (15) at (-0.75, 0.25) {};
		\node [style=none] (16) at (0.25, 0.25) {};
		\node [style={small_node}] (17) at (-0.25, 0) {};
		\node [style=none] (18) at (-4, 0) {};
		\node [style=none] (19) at (-3, 0) {};
		\node [style=none] (20) at (-3.5, 0.5) {};
		\node [style=none] (21) at (0.75, 0) {};
		\node [style=none] (22) at (1.75, 0.75) {};
		\node [style=none] (23) at (2.75, 0) {};
		\node [style=none] (24) at (3.25, 0) {};
		\node [style=none] (25) at (3.75, 0.5) {};
		\node [style=none] (26) at (4.25, 0) {};
		\node [style=none] (27) at (-3.5, 0) {};
		\node [style=none] (28) at (1.25, 0) {};
		\node [style=none] (29) at (1.75, 0) {};
		\node [style=none] (30) at (2.25, 0) {};
		\node [style=none] (31) at (3.75, 0) {};
		\node [style=none] (32) at (-1.25, -1.75) {};
		\node [style=none] (33) at (-0.25, -0.5) {};
	\end{pgfonlayer}
	\begin{pgfonlayer}{edgelayer}
		\draw [thick, bend left=90, looseness=2.00] (13.center) to (14.center);
		\draw [thick, in=180, out=-90, looseness=1.25] (14.center) to (12);
		\draw [thick, in=0, out=-90, looseness=1.25] (15.center) to (12);
		\draw [thick, bend left=90, looseness=1.75] (15.center) to (16.center);
		\draw [thick] (16.center) to (7.center);
		\draw [thick] (12) to (6.center);
		\draw [thick, bend right=90, looseness=1.25] (4.center) to (5.center);
		\draw [thick, bend right=90, looseness=1.25] (7.center) to (8.center);
		\draw [thick, bend right=90, looseness=1.00] (10.center) to (11.center);
		\draw [thick] (18.center) to (19.center);
		\draw [thick] (20.center) to (18.center);
		\draw [thick] (20.center) to (19.center);
		\draw [thick] (21.center) to (23.center);
		\draw [thick] (21.center) to (22.center);
		\draw [thick] (22.center) to (23.center);
		\draw [thick] (25.center) to (24.center);
		\draw [thick] (24.center) to (26.center);
		\draw [thick] (25.center) to (26.center);
		\draw [thick] (13.center) to (5.center);
		\draw [thick] (27.center) to (4.center);
		\draw [thick] (28.center) to (8.center);
		\draw [thick] (29.center) to (9.center);
		\draw [thick] (30.center) to (10.center);
		\draw [thick] (31.center) to (11.center);
		\draw [thick] (6.center) to (32.center);
		\draw [thick, bend left=90, looseness=1.25] (9.center) to (33.center);
		\draw [thick] (33.center) to (17);
	\end{pgfonlayer}
\end{tikzpicture}

%% file: tikz/frob-yanked.tikz
\begin{tikzpicture}[scale=0.55]
	\begin{pgfonlayer}{nodelayer}
		\node [style=none] (0) at (-1.75, 1.25) {\lang{Fruit}};
		\node [style=none] (1) at (0.25, 1.25) {\lang{tastes}};
		\node [style=none] (2) at (2.25, 1.25) {\lang{bitter}};
		\node [style=none] (3) at (-1.75, -0.5) {};
		\node [style=none] (4) at (-0.25, -0.5) {};
		\node [style=none] (5) at (0.25, -0.5) {};
		\node [style=none] (6) at (0.75, -0.5) {};
		\node [style=none] (7) at (2.25, -0.5) {};
		\node [style={small_node}] (8) at (0.25, -1.25) {};
		\node [style=none] (9) at (-2.25, 0) {};
		\node [style=none] (10) at (-1.25, 0) {};
		\node [style=none] (11) at (-1.75, 0.5) {};
		\node [style=none] (12) at (-0.75, 0) {};
		\node [style=none] (13) at (0.25, 0.75) {};
		\node [style=none] (14) at (1.25, 0) {};
		\node [style=none] (15) at (1.75, 0) {};
		\node [style=none] (16) at (2.25, 0.5) {};
		\node [style=none] (17) at (2.75, 0) {};
		\node [style=none] (18) at (-1.75, 0) {};
		\node [style=none] (19) at (-0.25, 0) {};
		\node [style=none] (20) at (0.25, 0) {};
		\node [style=none] (21) at (0.75, 0) {};
		\node [style=none] (22) at (2.25, 0) {};
		\node [style={small_node}] (23) at (-1, -1) {};
		\node [style=none] (24) at (-1, -1.75) {};
	\end{pgfonlayer}
	\begin{pgfonlayer}{edgelayer}
		\draw [thick] (5.center) to (8);
		\draw [thick, bend right=90, looseness=1.00] (6.center) to (7.center);
		\draw [thick] (9.center) to (10.center);
		\draw [thick] (11.center) to (9.center);
		\draw [thick] (11.center) to (10.center);
		\draw [thick] (12.center) to (14.center);
		\draw [thick] (12.center) to (13.center);
		\draw [thick] (13.center) to (14.center);
		\draw [thick] (16.center) to (15.center);
		\draw [thick] (15.center) to (17.center);
		\draw [thick] (16.center) to (17.center);
		\draw [thick] (18.center) to (3.center);
		\draw [thick] (19.center) to (4.center);
		\draw [thick] (20.center) to (5.center);
		\draw [thick] (21.center) to (6.center);
		\draw [thick] (22.center) to (7.center);
		\draw [thick, bend right=45, looseness=1.00] (3.center) to (23);
		\draw [thick, bend left=45, looseness=1.00] (4.center) to (23);
		\draw [thick] (23) to (24.center);
	\end{pgfonlayer}
\end{tikzpicture}